\documentclass[letterpaper, 10 pt, conference]{ieeeconf}  

\IEEEoverridecommandlockouts                              

\usepackage{graphics} 
\usepackage{epsfig} 
\usepackage{mathptmx} 
\usepackage{times} 
\usepackage{amsmath} 
\usepackage{amssymb}  

\title{\LARGE \bf
Evaluation of an Actuated Spine in Agile Quadruped Locomotion
}

\author{Nico Bohlinger$^{1}$*, Piotr Kicki$^{2}$*, Davide Tateo$^{1}$, Krzysztof Walas$^{2}$ and Jan Peters$^{1}$
\thanks{$^{1}$The authors are with the Department of Computer Science, Technical University of Darmstadt, 64289 Darmstadt, Germany.
}
\thanks{$^{2}$ The authors are with IDEAS Research Institute, 00-060 Warsaw, Poland, and the Institute of Robotics and Machine Intelligence, Poznan University of Technology, 60-965 Poznan, Poland
}
\thanks{* Denotes equal contribution}
\thanks{
{Corresponding author: \tt\small piotr.kicki@put.poznan.pl}%
}}

\begin{document}

\maketitle
\thispagestyle{empty}
\pagestyle{empty}

\begin{abstract}
The spine plays a crucial role in the dynamic locomotion of quadrupedal animals, improving the stability, speed, and efficiency of their gait, especially for fast-paced and highly agile movements \cite{hildebrand1959motions, schilling2006sagittal}.
Therefore, the spine is also a promising and natural way to extend the capabilities of quadruped robots \cite{khoramshahi2013benefits, hyun2014high, matsumoto2023high, li2023dynamic, yoneda2025kleiyn}.
This paper empirically investigates the benefits of an actuated spine for learning agile quadruped locomotion. We evaluate whether the use of the spine brings benefits in terms of high-speed running, climbing stairs, climbing high-angle slopes, hurdling, and crawling scenarios.
We conducted an empirical study in MuJoCo simulation using the Silver Badger robot from MAB Robotics with an actuated 1-DOF spine in the sagittal plane. The obtained results show that the use of the spine provides the robot with increased agility and allows it to overcome higher stairs, steeper slopes, higher obstacles, and smaller passages.
\end{abstract}

\vspace{1em}

\section{Empirical study}

To empirically test whether the use of an actuated spine provides some advantages over the robot without it, we conducted a set of experiments in the MuJoCo simulator \cite{todorov2012mujoco} with the Silver Badger robot from MAB Robotics with an actuated 1-DOF spine in the sagittal plane.
In particular, motivated by the recent advancements in reinforcement learning-based agile quadruped locomotion~\cite{Tan-RSS-18,parkour, rudin2025parkourwildlearninggeneral,bohlinger2024onepolicy}, we perform this comparison with policies trained using Proximal Policy Optimization \cite{schulman2017proximal} implemented with RL-X \cite{bohlinger2023rlx}. 
In all experiments, we use domain randomization and add observation noise to facilitate transferability to the real robot.


\subsection{High-speed running}

The evaluation of agile locomotion needs effective metrics for locomotion speed and energy efficiency, allowing for a fair comparison between robots of different types and sizes.
The Froude number \cite{alexander1984gaits} measures the size-independent locomotion speed, and the cost of transport \cite{tucker1975energetic} quantifies the energy efficiency of the locomotion.
The two dimensionless metrics are defined as follows: 

\small
$$
\text{Fr} = \frac{v^2}{gh}, \quad \text{COT} = \frac{P}{mgv}
$$
\normalsize
where $v$ is the forward velocity, $g$ is the gravitational acceleration, $h$ is the height of the hips, $P$ is the used power and $m$ is the mass.

To investigate the robot's energy efficiency and maximum running velocity, we designed an automatic learning curriculum that gradually increases the target velocity.
Fig. \ref{fig:highspeed} shows that, with a locked spine, the robot achieves a maximum of 5.0 m/s forward velocity at the end of training.
Enabling the spine during learning improves the maximum velocity to 5.9 m/s.
This increase in velocity is also reflected in the Froude number, and an improvement in stability can be seen in the average episode length.
The active spine achieves a $\text{Fr} \approx 9.7$ while the locked spine only reaches a $\text{Fr} \approx 6.7$ with a 3\% shorter episode length.
Visual inspection shows that the RL policy learns to contract the body with the spine to pull the legs forward and prepare the next step, which leads to a more natural-looking, faster, and stable gait.
When utilizing the motor of the spine joint, the energy consumption increases by 6.2\% (81 Wh) at the maximum running velocity, but through the higher increase in top speed, this leads to a more energy-efficient gait at the end with a $\text{COT} \approx 2.1$ compared to the locked spine with a $\text{COT} \approx 2.3$.

\begin{figure}[t]
\centering
\includegraphics[scale=0.16]{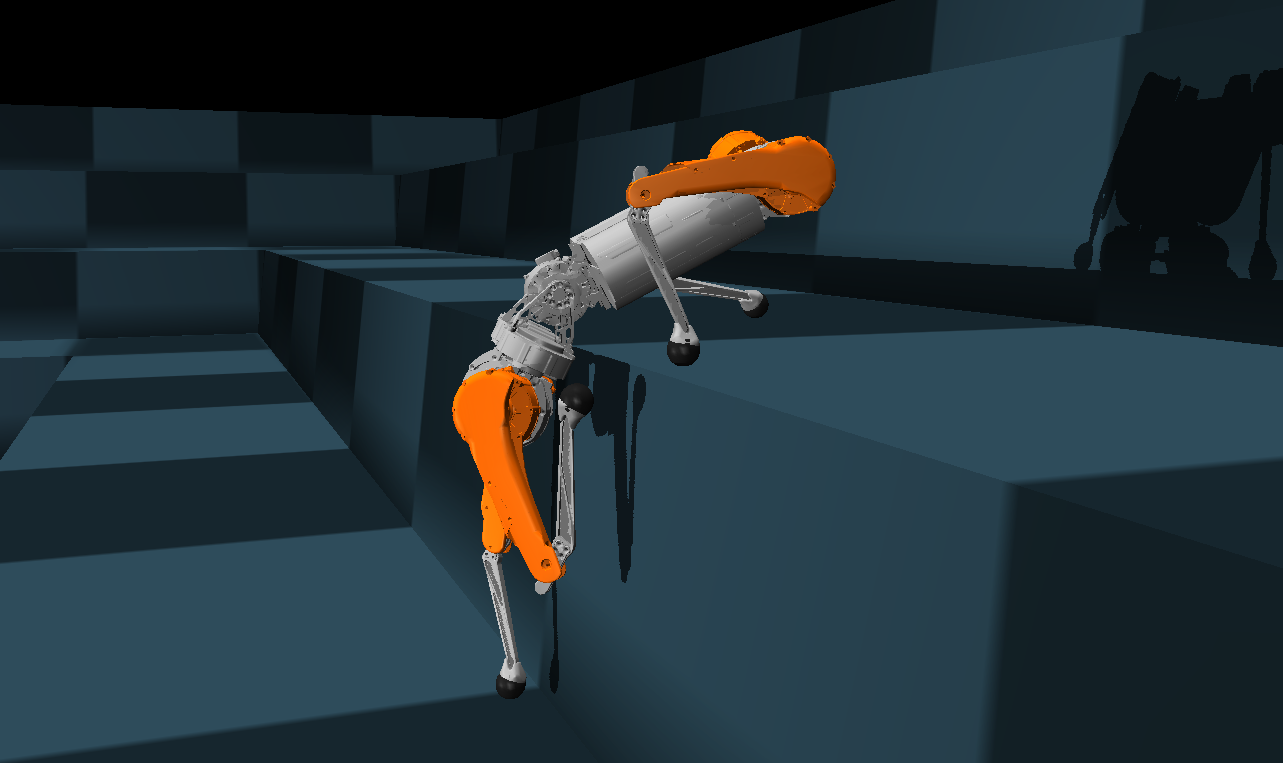}
\caption{Inverted pyramid stairs for the climbing experiments in simulation. The robot actively uses its spine to bend the front part of the trunk over the stairs and get leverage for the rest of its body.}
\label{figure2}
\vspace{-1em}
\end{figure}

\begin{figure}[b]
\vspace{-1em}
\centering
\includegraphics[width=0.9\linewidth]{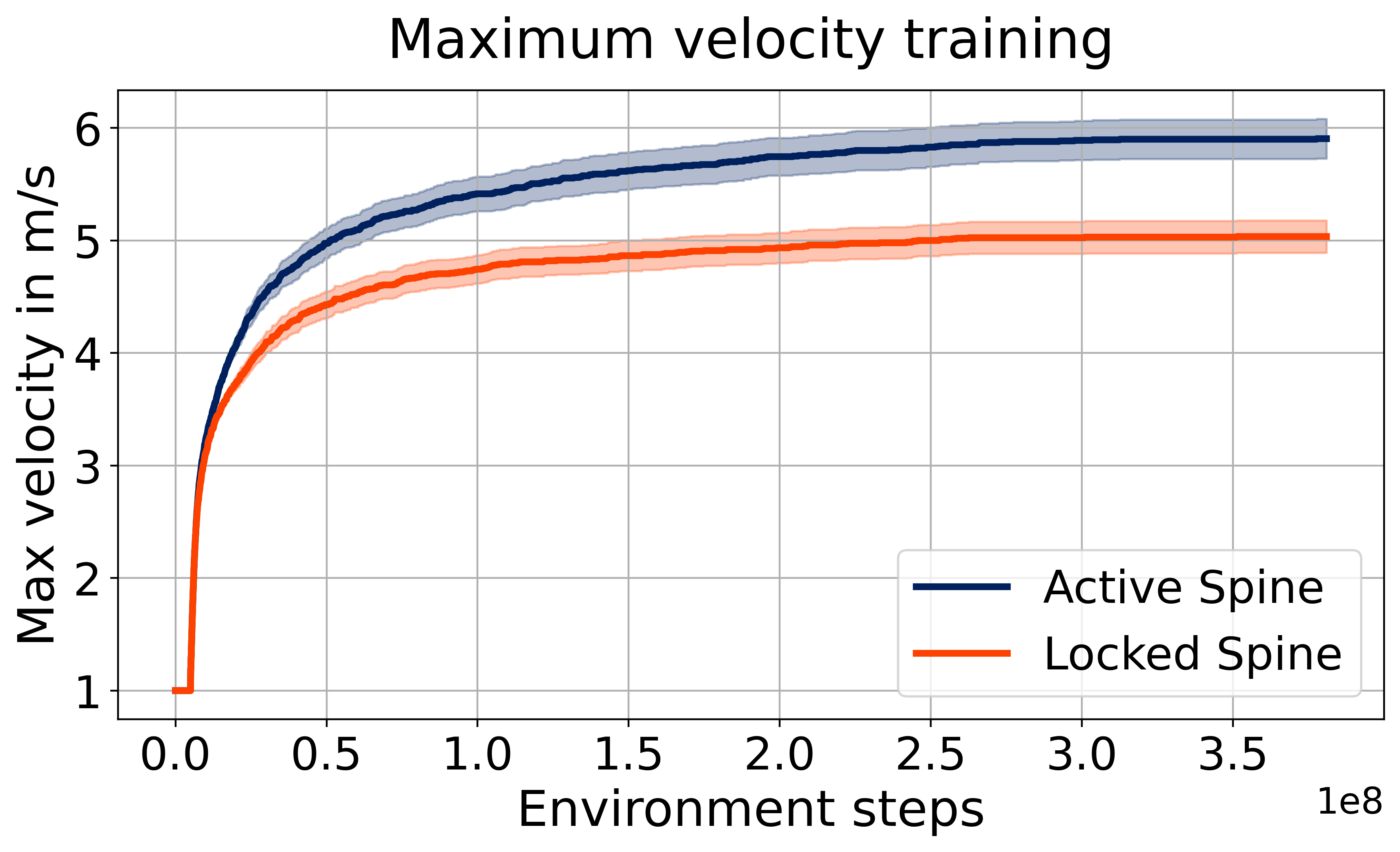}
\caption{The maximum forward velocity achieved by the robot during learning. The thick line and shaded area show the mean and the standard error, respectively, over 20 seeds.}
\label{fig:highspeed}
\end{figure}

\subsection{Climbing}

Besides fast running, another task that requires a high level of agility is climbing.
To identify the climbing capabilities of the Silver Badger, we created an inverted pyramid environment with evenly spaced stairs (Fig. \ref{figure2}).
Again, an automatic curriculum guides the learning process by increasing the height of the stairs every time the robot successfully climbs seven consecutive stairs.
At the end of training, the maximum stair height the robot can climb is 0.7 m, which is 2.2x its standing height.
On average, the version with the active spine can climb 4 cm higher while consuming the same amount of energy as the robot with the locked spine.

\begin{figure}[t]
\vspace{-1em}
\centering
\includegraphics[width=0.9\linewidth]{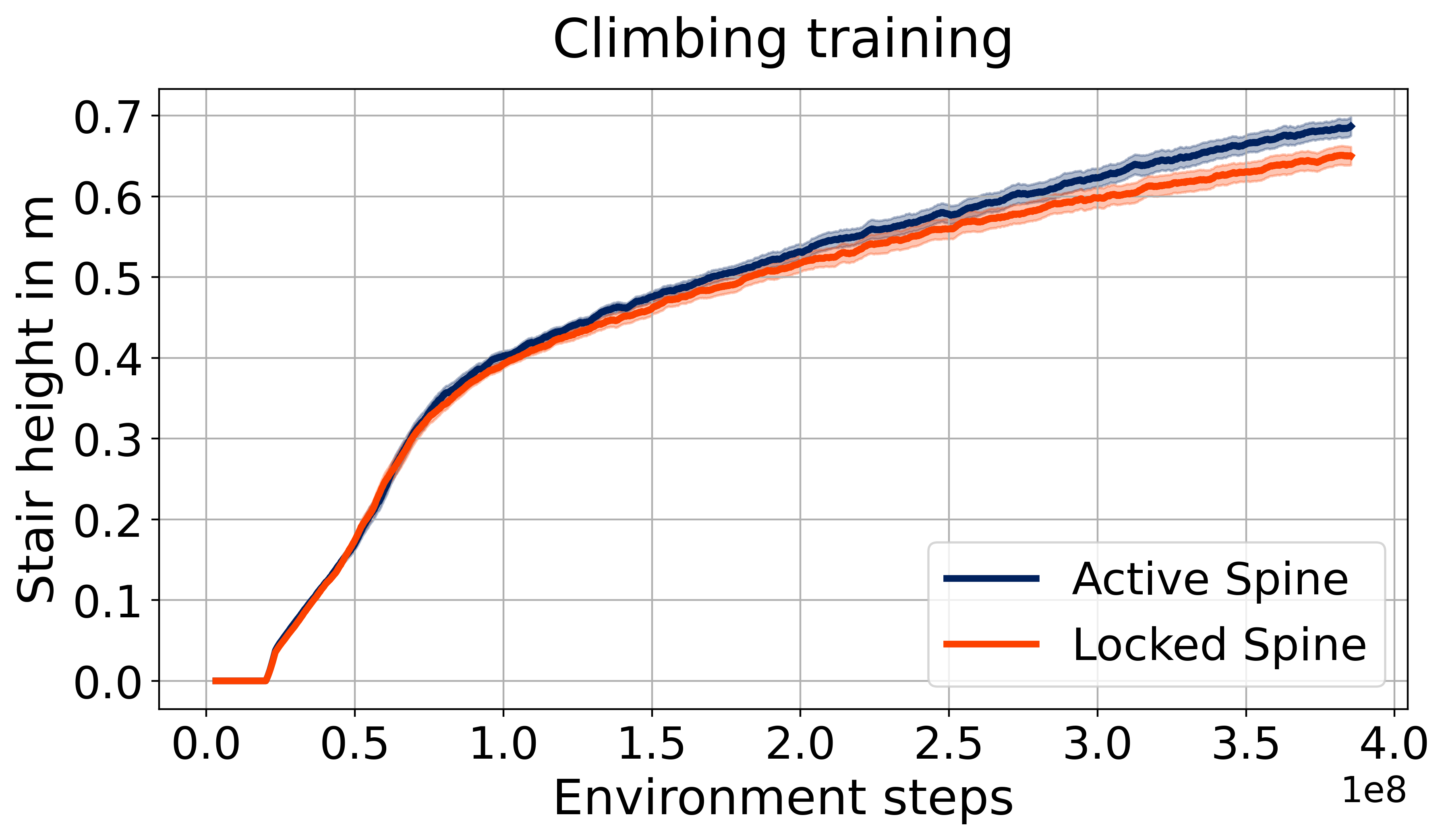}
\caption{The maximum stair height climbed by the robot during learning. The thick line and shaded area show the mean and the standard error, respectively, over 20 seeds.}
\label{fig:stairs}
\end{figure}

\begin{figure}[t]
\vspace{-1em}
\centering
\includegraphics[width=0.9\linewidth]{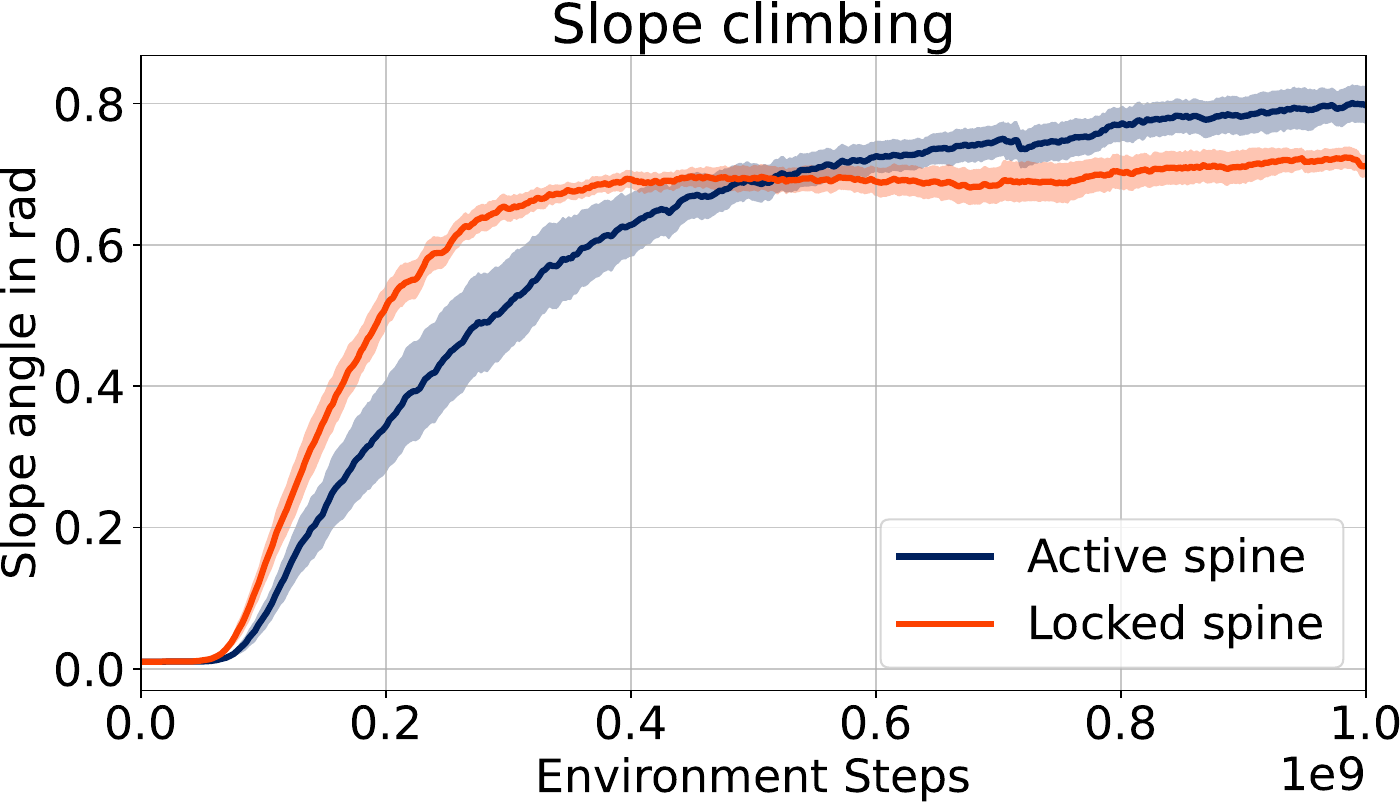}
\caption{The maximum slope angle climbed by the robot during learning. The thick line and shaded area show the mean and the standard error, respectively, over 15 seeds.}
\label{fig:ramp}
\end{figure}

Another climbing scenario in which we evaluated the actuated spine is slope climbing. In this case, we created a 4 m by 4 m environment with a 1 m wide slope for the robot to climb. Also, here, an automatic curriculum guides the learning process by increasing the slope angle by 0.01 rad every time the robot reaches the top of the slope and decreasing the slope angle if the robot fails to reach the halfway point. The progress of the maximum slope angle during training is presented in Fig.~\ref{fig:ramp}. One can see that the robot with the locked spine learns faster in the first phase, but after reaching angles of about 0.7 rad, its performance stabilizes. Instead, a robot with an active spine needs more time to reach the same performance, but then steadily progresses towards slopes of 0.8 rad on average.

\subsection{Overcoming obstacles}

\begin{figure}[b]
\vspace{-1em}
\centering
\includegraphics[width=0.85\linewidth]{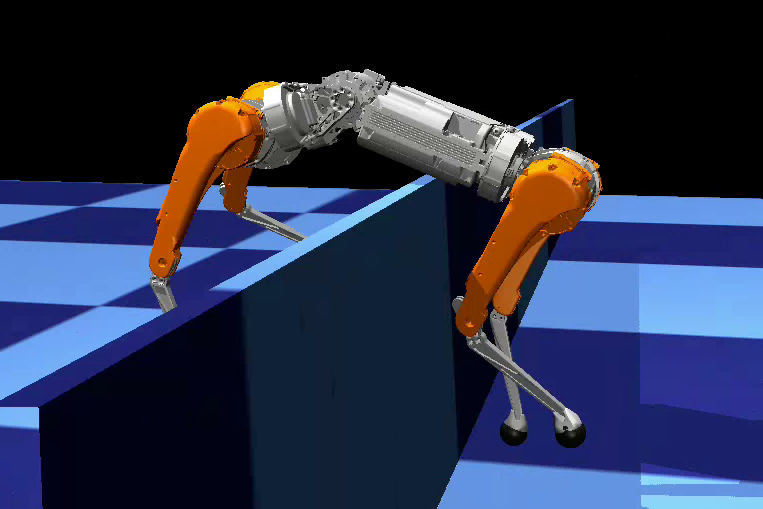}
\caption{Hurdling scenario, in which the robot is tasked to overcome the vertical obstacle located on the ground. The robot actively uses its spine to increase the clearance while jumping over the wall.}
\label{fig:wall_jump}
\end{figure}

\begin{figure}[t]
\vspace{-1em}
\centering
\includegraphics[width=0.85\linewidth]{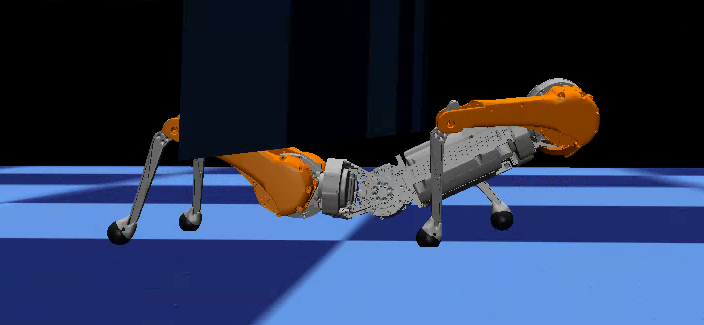}
\caption{Crawling scenario, in which the robot is tasked to overcome the vertical obstacle located on the ceiling. The robot actively uses its spine to increase the distance to the obstacle while crawling beneath it.}
\label{fig:crawl}
\end{figure}

One of the key features of agile locomotion is the ability to overcome obstacles. To investigate the usefulness of the actuated spine in overcoming obstacles, we designed two environments that contain a single vertical obstacle. In the first scenario -- hurdling, the obstacle is a 5 cm thick wall located on the ground (see Fig.~\ref{fig:wall_jump}), whereas in the second scenario -- crawling, an analogous wall is placed at the ceiling (see Fig.~\ref{fig:crawl}). Similarly, like in the previous environments, we implemented an automatic curriculum that raises the height of the wall in the first scenario, and reduces the height of the passage by increasing the height of the ceiling wall in the second one. Every time the robot overcomes the currently highest obstacle, its height increases by 5 mm, whereas when it fails to do so, there is a slight chance that its height will decrease by 5 mm.

\begin{figure}[t]
\vspace{-1em}
\centering
\includegraphics[width=0.9\linewidth]{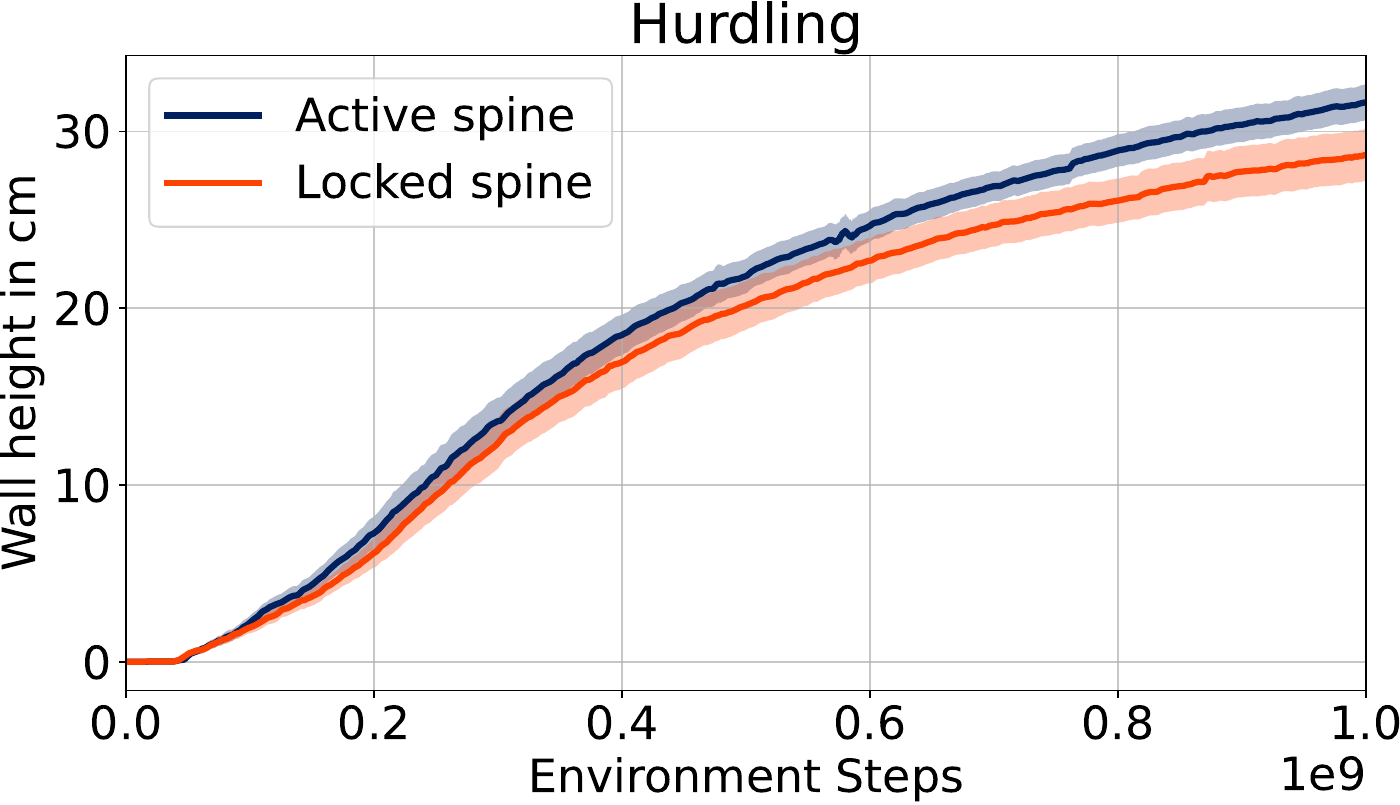}
\caption{The maximum wall height overcome by the robot during learning. The thick line and shaded area show the mean and the standard error, respectively, over 15 seeds.}
\label{fig:wall}
\end{figure}

\begin{figure}[t]
\vspace{-1em}
\centering
\includegraphics[width=0.9\linewidth]{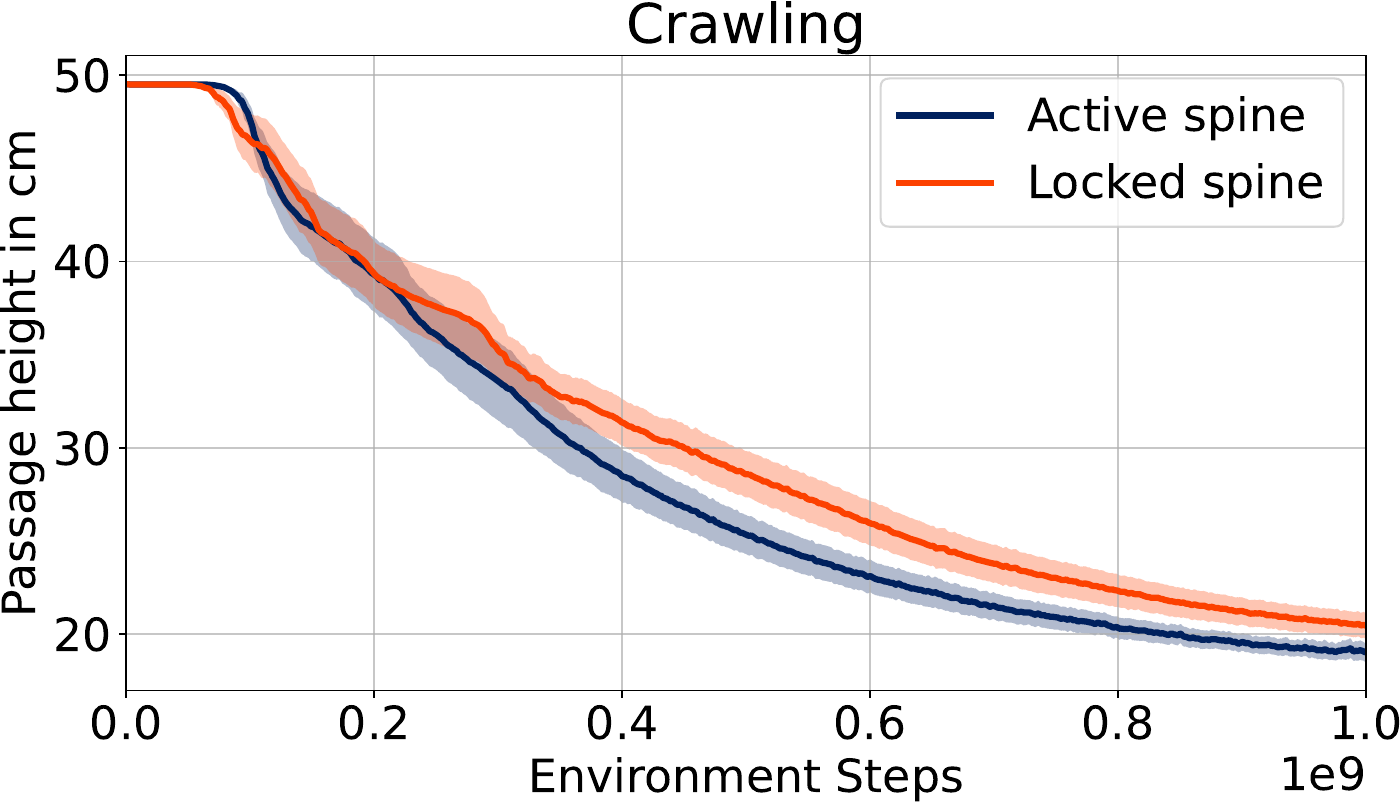}
\caption{The minimum height of the passage overcome by the robot during learning. The thick line and shaded area show the mean and the standard error, respectively, over 15 seeds.}
\label{fig:wallceiling}
\end{figure}

The resulting evolutions of the obstacle and passage heights during training are presented in Fig.~\ref{fig:wall} and Fig.~\ref{fig:wallceiling}, for hurdling and crawling, respectively.
One can see that in both cases, the use of the spine proves to be beneficial in overcoming obstacles, as it allows one to jump over 4 cm taller walls and crawl through 2 cm smaller passages, on average.

\section{Conclusions}
This empirical study shows the benefits of the actuated spine in increasing the agility of a quadruped robot, especially in the tasks of high-speed locomotion, climbing stairs and slopes, and overcoming vertical obstacles.





\section*{ACKNOWLEDGMENTS}
This project was funded by the National Science Centre, Poland, under the OPUS call in the Weave programme UMO-2021/43/I/ST6/02711, and by the German Science Foundation (DFG) under
grant number PE 2315/17-1. The work of Piotr Kicki was supported by the Foundation for Polish Science (FNP).
The authors gratefully acknowledge the computing time provided to them on the high-performance computer Lichtenberg at the NHR Centers NHR4CES at TU Darmstadt and on the infrastructure of the Poznan Supercomputing and Networking Center.

\newpage

\bibliographystyle{IEEEtran}
\bibliography{IEEEabrv,bib}

\end{document}